\documentclass[letterpaper, 10 pt, conference]{ieeeconf}  

\IEEEoverridecommandlockouts                              

\usepackage{graphicx} 
\usepackage{amsmath} 
\usepackage{amssymb}  
\usepackage{multirow}
\usepackage{xspace}
\usepackage{subcaption}

\makeatletter
\DeclareRobustCommand\onedot{\futurelet\@let@token\@onedot}
\def\@onedot{\ifx\@let@token.\else.\null\fi\xspace}
\def\eg{\emph{e.g}\onedot} 
\def\ie{\emph{i.e}\onedot} 
 
\def\etc{\emph{etc}\onedot}

\makeatother

\title{\LARGE \bf
Self-Supervised Road Layout Parsing with Graph Auto-Encoding
}

\author{Chenyang Lu and Gijs Dubbelman
	\thanks{
		This work was supported by the Netherlands Organization for Scientific Research (NWO) in the context of the i-CAVE project.}
	\thanks{Authors are with the Mobile Perception Systems research lab of the SPS/VCA group, Department of Electrical Engineering,  Eindhoven University of Technology, The Netherlands.
		{\tt\footnotesize \{c.lu.2, g.dubbelman\}@tue.nl}}%
}

\begin{document}

\maketitle
\thispagestyle{empty}
\pagestyle{empty}


\begin{abstract}

Aiming for higher-level scene understanding, this work presents a neural network approach that takes a road-layout map in bird's-eye-view as input, and predicts a human-interpretable graph that represents the road's topological layout. Our approach elevates the understanding of road layouts from pixel level to the level of graphs. To achieve this goal, an image-graph-image auto-encoder is utilized. The network is designed to learn to regress the graph representation at its auto-encoder bottleneck. This learning is self-supervised by an image reconstruction loss, without needing any external manual annotations. We create a synthetic dataset containing common road layout patterns and use it for training of the auto-encoder in addition to the real-world Argoverse dataset. By using this additional synthetic dataset, which conceptually captures human knowledge of road layouts and makes this available to the network for training, we are able to stabilize and further improve the performance of topological road layout understanding on the real-world Argoverse dataset. The evaluation shows that our approach exhibits comparable performance to a strong fully-supervised baseline.

\end{abstract}

\section{Introduction}

\begin{figure*}[!tbp]
	\centering
	\includegraphics[width=0.95\linewidth]{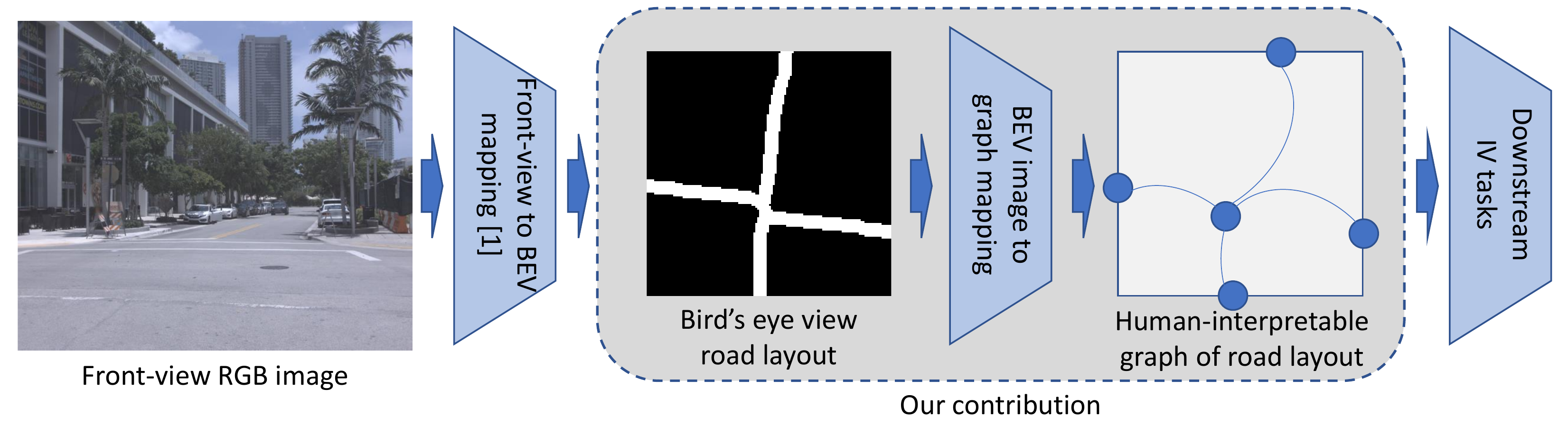}
	\caption{Intelligent vehicles are able to capture rich photometric data at low cost, and obtain semantic understanding of scenes using state-of-the-art artifical intelligence techniques, \eg, transforming front-view RGB images into bird's-eye-view (BEV) road layouts~\cite{yang_projecting_2021}. In this work, we make one step forward towards high-level understanding of urban scenes for intelligent vehicles: parsing road layouts as human-interpretable graphs. As visualized above, given a BEV road layout as input, our neural network is able to predict a graph describing the road layout scene, which contains \textit{road joints} as \textit{nodes}, and their \textit{connectivity} as \textit{edges}. Training of our network is self-supervised and thus does not require any manual annotation of nodes and edges, which makes our approach more scalable than fully-supervised methods.}
	\label{fig_overview}
\end{figure*}

Scene understanding is a crucial upstream task for intelligent vehicles, which is relied on by various downstream intelligent vehicle tasks, such as decision making and route planning. With the developments of deep convolutional neural networks (CNNs), significant advances have been made for scene understanding: from semantic segmentation \cite{shelhamer_fully_2017} to object detection \cite{he_mask_2017}, which are widely deployed to realize automated driving functions. In this work, to further improve scene understanding capabilities of intelligent vehicles, we aim to extract more high-level scene information from front-view images, \eg, topological road layouts represented by graphs.

Current approaches that understand environments, as discussed in Section~\ref{sect_related_work}, usually result in pixel-related representations, \eg, pixel-wise labels, bounding boxes with pixel coordinates, \etc. They are common and widely used in the computer vision community, while for intelligent vehicle applications, many downstream tasks typically do not act on pixel-level information but require more higher-level abstract representations, \eg, digital graph/vector maps. For instance, given an image depicting an urban scene, an off-the-shelf CNN can segment out the road pixels seamlessly \cite{shelhamer_fully_2017}, or even perform bird's-eye-view (BEV) transformation simultaneously \cite{yang_projecting_2021}, see the left part of Figure~\ref{fig_overview}. However, these approaches lack the inherent structural understanding of the road, such as its topology, which may be preferred by downstream decision making and route planning tasks. 

Recently, a scene graph representation \cite{johnson_image_2015} was proposed where scene understanding capabilities are enriched by assigning relationship labels between detected objects that are semantically associated. Inspired by this, we aim to bring pixel-level semantic understanding of road layouts to a higher structural level, see Figure~\ref{fig_overview}. Specifically, given a front-view RGB image containing urban scenes, the end-to-end task starts with estimating a road layout map in BEV, and then parsing this map as a graph. The graph parsing task is to extract the \textit{road joints} as a set of \textit{nodes}, and further predict the \textit{connectivity} between arbitrary two detected nodes as \textit{edges}. Together, the set of nodes and edges, compose a \textit{graph} that describes the observed road layout in an efficient and structural manner. Since various approaches have been proposed to perform front-view to BEV road layout mapping \cite{lu_monocular_2019, mani_monolayout_2020, yang_projecting_2021}, in this work we focus on the graph parsing task from BEV road layouts, which is highlighted as the second stage in Figure~\ref{fig_overview}. The alternative of using model-based approaches, \eg, hand-crafted feature extractors and designed image processing logic, to parse the BEV road layouts into a graph currently offers a practical solution. However, the potential scalability and accuracy of deep data-driven approaches are much higher than that of model-based designs. Neural networks have the potential to deal with various semantic labels for joint and connectivity, as in OpenStreetMap \cite{haklay_openstreetmap:_2008}, unlike hand-crafted approaches that cannot scale beyond the requirements for which they have been designed. Therefore, to potentially realize more scalable and better performing graph-parsing models, in this research, we explore a novel deep data-driven design to perform the graph parsing task.

A significant challenge exists when deploying a data-driven approach to perform the aforementioned task: although various state-of-the-art network architectures exist \cite{xu_scene_2017, herzig_mapping_2018, zellers_neural_2018, suhail_energy-based_2021}, almost all of them require massive annotated ground truth for training. Despite the outstanding performance they achieve on public benchmarks \cite{krishna_visual_2017}, it is not possible to apply them onto a novel domain-specific application, such as parsing road layouts as graphs, unless one manually annotates massive samples with road joints and their corresponding connectivity edges. To avoid the massive annotation efforts for fully-supervised deep learning, which also lacks scalability, we explore and extend the self-supervised image-graph-image auto-encoder methodology that is recently proposed in more fundamental work \cite{lu_image-graph-image_2020}. In \cite{lu_image-graph-image_2020}, graphs are learned without needing manual annotations via auto-encoding from a synthetic dataset containing only simple shapes that have no applicability to real-world tasks. In this work, which is detailed in Section~\ref{sect_method}, we extend this approach to real-world road layouts.

We perform a series of experiments using real-world Argoverse \cite{chang_argoverse_2019} dataset, as described in Section~\ref{sect_exp}. To evaluate the quality of predicted graphs, we manually annotate 500 Argoverse samples based on their topological patterns, and evaluate the prediction performance using road topology classification accuracy. The results in Section~\ref{sect_results} show that our approach exhibits decent performance compared to a strong fully-supervised baseline, and has several use cases for real applications.

The contributions of our work can be summarized as:
\begin{itemize}
	\item We generalize the previously proposed image-graph-image auto-encoder \cite{lu_image-graph-image_2020} to real-world road layouts;
	\item We showcase two use-cases of predicting graph representation of road layouts, including a pipeline from front-view RGB to road layout graphs, and large-scale road layout parsing;
	\item We make our code and manually annotated road topology test labels available to the community \cite{linktocode}.
\end{itemize}

\section{Related work}
\label{sect_related_work}

\textbf{Pixel-based scene understanding} Scene understanding is vital for intelligent vehicles, which has also been a broad and challenging topic in the computer vision community. Various sub-tasks with corresponding solutions are proposed step-by-step towards a more comprehensive understanding of images: from object classification \cite{krizhevsky_imagenet_2012} to detection \cite{girshick_rich_2014}, and from semantic segmentation \cite{shelhamer_fully_2017} to part-aware panoptic segmentation \cite{de_geus_part-aware_2021}. Each evolutionary step introduced extra information that a network could deliver, and took a step forward towards a more holistic understanding of scenes. Despite the achieved advances, the understanding capability of machines is still distant from that of humans in terms of robustness, generalization, and efficiency. One potential reason is that the processed representations by machines often remain at lower abstraction level, (\eg, per-pixel labels, and bound-box coordinates), which are not well suited for higher-level automated reasoning on the observed scenes. Thus, it is preferred to process and represent the scene using a more structural, efficient, and informative format instead of image-like tensors, such as scene graph \cite{johnson_image_2015}.

The higher-level representation could be generated by rule-based or model-based methods. For instance, one could perform traditional approaches to find out the road joints, and then classify the connectivity between them. However, these approaches often lack scalability and generalizability. Thus, in this work we attempt to tackle this task under the paradigm of deep learning, since in our case, neural networks have the potential to deal with various semantic labels for key joint and connectivity in future developments.

\textbf{Scene graphs} With the aforementioned goal, scene graphs, were pioneered by \cite{johnson_image_2015}, where detected objects from input images are modeled as nodes and semantic relations between them are modeled as edges with semantic relational labels. Recently, more approaches \cite{xu_scene_2017, li_factorizable_2018, qi_attentive_2019, liu_fully_2021} are proposed on top of this and deliver state-of-the-art performance and efficiency on public benchmarks like Visual Genome \cite{krishna_visual_2017}. Naturally, all of them rely on ground truth provided by the dataset for training deep scene graph generation models, and this holds for each new domain-specific task. The key difference between these state-of-the-art methods and our method is that, we do not rely on the massive annotated ground truth for training, and the graphs are learned in an end-to-end self-supervised manner. Although our approach is currently restricted to parse binary road layouts, we believe it brings the benefits of deep learning approaches to this task, which is currently often pragmatically solved using rule-based or model-based solutions. 

\textbf{Road parsing and modeling} Closer to our approach from an application perspective, representations and corresponding processing approaches have been proposed to describe the road layout, which directly benefit scene understanding for intelligent vehicles. One direction is to parse raw sensory input (\eg, images from front-view camera) into BEV, which provides a straightforward map of the local road layout, as proposed in \cite{lu_monocular_2019}. Further work improved the quality of road layout and especially extend the unobserved region due to the sensor's field-of-view limitation, as post-processing \cite{lu_learning_2020, schulter_learning_2018} or part of end-to-end solution \cite{mani_monolayout_2020, yang_projecting_2021}. On top of this, another road layout model \cite{wang_parametric_2019} was proposed, which performs a similar task but provides an extended road scene model with enriched elements. Research has also been carried out on representing road layouts in a more structural manner instead of taking them as an image-like tensor or a semantic occupancy grid map. In \cite{kunze_reading_2018}, a scene graph is extracted based on the segmented road elements (\eg, road markings, curbs, \etc) using a rule-based algorithm as post-processing. Our work is close to \cite{kunze_reading_2018} in terms of the predicted graph representations, but is achieved using self-supervised deep learning without pre-defined rules and models.

\section{Methodology}
\label{sect_method}

\begin{figure*}[!tbp]
	\centering
	\includegraphics[width=\linewidth]{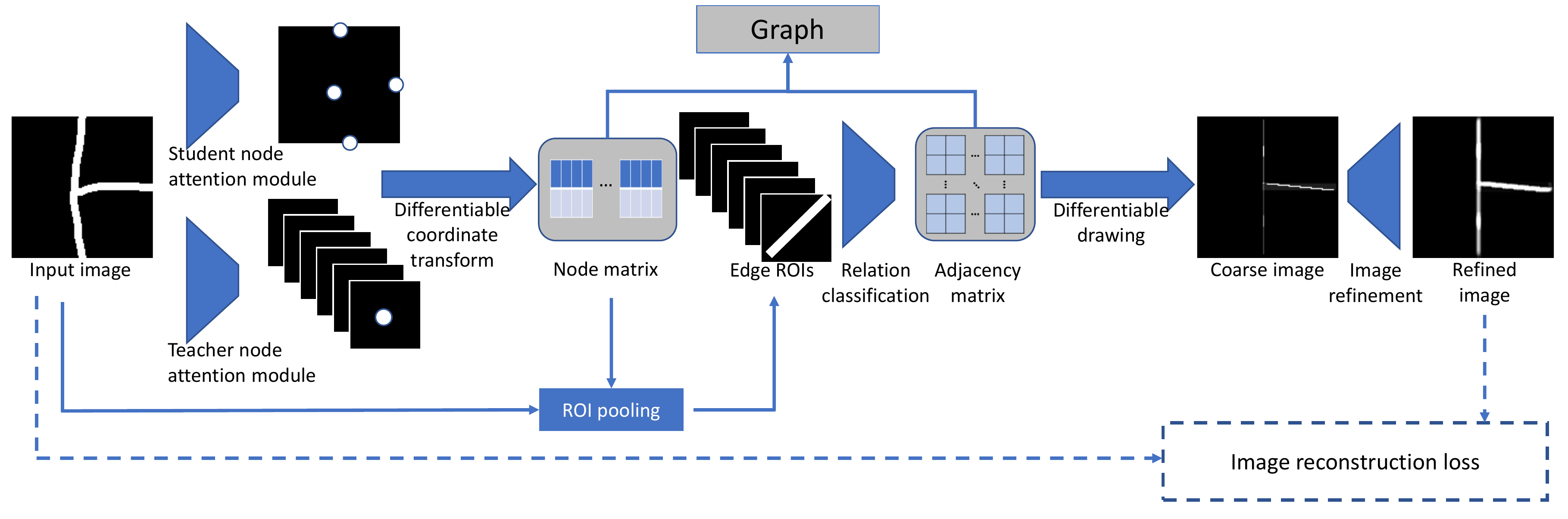}
	\caption{Network structure. The design is from the previous proposed approach for simple shape parsing as graphs \cite{lu_image-graph-image_2020}. In this case, the input images for self-supervised learning are real-world road layouts together with the synthetic road layouts. Please see Section~\ref{sect_exp} for details.}
	\label{fig_network}
\end{figure*}

\subsection{Problem definition}
\label{sect_method_pro_def}

As illustrated in Figure~\ref{fig_overview}, given a front-view image $I_{\text{front-view}}$ that contains the road layout, we assume there exists a graph $G = (V, E)$ that describes the road layout in a human-interpretable and structural manner. To be specific, $V$ is a set of positional coordinates of \textit{road joints} $v_i$; and $E$ is an adjacency matrix with each element $e_{ij}$ representing the connectivity between joint $v_i$ and joint $v_j$. The graph $G$ contains all the key information one needs to understand the road layout, and one is able to explain the physical and semantic meaning of every element. The proposed data definition is inspired by contemporary geographic map projects, \eg, OpenStreetMap \cite{haklay_openstreetmap:_2008}, where the database also contains nodes and edges representing the road layout. Each node, carrying its geographical location and a unique identifier number, represents one single point on the map; and each edge represents a poly-line on the map, carrying and connecting an ordered list of node identifiers. Our proposed data definition is similar to these commonly used data formats, which are already widely used for modern navigation applications, thus it could be directly used by various downstream tasks. 

With the defined output format, the goal of our approach is to predict the aforementioned graphs from front-view RGB images. We decompose the pipeline into two stages, with the first stage being the mapping from front-view RGB image to BEV binary road layouts, which is realized by one of the state-of-the-art networks \cite{yang_projecting_2021}. We denote the first stage as 
\begin{equation}
	I_{\text{BEV}} = f_{pre}(I_{\text{front-view}}),
\end{equation}
where $I_{\text{BEV}}$ is the binary road layout image, and $f_{pre}$ is the off-the-shelf mapping approach \cite{yang_projecting_2021}.

Afterward, we take the binary road layout image as input and process it using our proposed network, which is formalized as
\begin{equation}
	G = f(I_{\text{BEV}}),
\end{equation}
where $f$ is the mapping function that takes BEV road layout $I_{\text{BEV}}$ as input and predicts graph $G$. This second stage is the key task to focus in this work, and the complete two-stage pipeline is presented later as an application experiment in Section~\ref{sect_resluts_full_pipeline}.

\subsection{Network-based approach}

We adopt the auto-encoder network developed in \cite{lu_image-graph-image_2020} and deploy it to realize the aforementioned task, as illustrated in Figure~\ref{fig_network}. It is initially used for leaning graphs from simple synthetic shapes \cite{lu_image-graph-image_2020}, and in this work we extend it to handle real-world road layouts with extra training configurations, as detailed in Section~\ref{sect_method_training}. The network is composed of an encoder $f$ and a decoder $g$. While the encoder $f: I_{\text{BEV}} \rightarrow G$ learns to predict graph from input road layout and is solely used for inference, the decoder $g: G \rightarrow I_{\text{BEV}}$ learns to take the graph predictions from $f$ to reconstruct the input road layout as image, which is only required during training for passing gradients  to the encoder $f$. By training the auto-encoder with image reconstruction loss, the graph information is learned without any external supervision. The complete learning pipeline can be formalized as 
\begin{equation}
\begin{split}
	\tilde{G} = f(I_{\text{BEV}})\\
	\tilde{I}_{\text{BEV}} = g(\tilde{G}).
\end{split}
\end{equation}

The following paragraphs briefly introduce the design and functionality of our proposed network, and please see \cite{lu_image-graph-image_2020} for more details and \cite{linktocode} for access to our implementation.

\textbf{Encoder} The encoder $f$ is designed to perform two sub-tasks sequentially: 1) road joint detection, and 2) joint connectivity classification, which is illustrated at the left part of Figure~\ref{fig_network}. Given an input road layout image, two conventional CNN-based modules, \ie, \textit{teacher node attention module} and \textit{student node attention module}, are first applied to detect the joints of road layouts. During training, the teacher module learns to attend an interested joint for each channel in the output image-like tensor. Afterward, the attended joints are transformed into image coordinates in a differentiable manner, which is used to generate a set of local regions-of-interest (ROIs) between arbitrary two detected joints using ROI pooling. These ROIs are taken as inputs by a CNN-based relation classifier that estimates the binary connectivity between arbitrary two joints. Together, the predicted \textit{node matrix} containing coordinates and \textit{adjacency matrix} containing connectivities compose a complete \textit{graph} for each input road layout. 

Please note that along with the teacher node attention module, we deploy a \textit{student node attention module} for the same node attention task. The difference is that the student module has no pre-defined number of nodes, \ie, fixed number of channels in output tensor, as in the teacher module, and attends all the road joints on a single channel. By doing so, the student module is able to generalize to images with arbitrary number of road joints, but cannot be fully differentiable during training. Thus, it learns to attend joints by the supervision from the teacher module, which itself is purely trained in a self-supervised manner. \textit{Only} the student module is used \textit{during inference} in replacement of the teacher module.

\textbf{Decoder} The decoder, visualized at the right part of Figure~\ref{fig_network}, is used to pass the supervision signal from image reconstruction loss to the encoder during training. It also performs another two sub-tasks: 1) differentiable drawing and 2) image refinement. The differential drawing module is created based on the gird sampling technique from spatial transformer network \cite{jaderberg_spatial_2015}. It takes the graph from the encoder as input and draws road segments onto a blank image in a differentiable manner. To be specific, one can convert a road graph into a set of triplets, each containing a pair of connected road joints with coordinates. Our differential drawing module can take each triplet as input and copy-paste a template road segment onto the blank image, which results in a coarse reconstructed road layout image. Afterward, an image refinement module is applied to merge the domain gap between coarse reconstructed images and input images and predict the refined images for self-supervised learning. The key here is that this whole drawing operation is differentiable such that the gradients from the image reconstruction loss can flow through the decoder to update the encoder.

\subsection{Training with real and synthetic data}
\label{sect_method_training}

With the encoder and decoder sequentially connected, the complete auto-encoder is trained with road layout images in a self-supervised manner. We simply feed images to the proposed auto-encoder, and encourage it to reconstruct the input image at the output of the decoder, by applying similarity loss between reconstructed and input images. We use the same loss function $\mathcal{L}$ as in \cite{lu_image-graph-image_2020}, \ie, multi-scale structural similarity index (SSIM). The training procedure can be formalized as 
\begin{equation}
	\begin{array}{l}
		f,g = \underset{f,g}{\arg\min}\ \mathcal{L}(I_{\text{BEV}}, g(f(I_{\text{BEV}}))), \\
	\end{array}
\end{equation}
where $I_{\text{BEV}}$ is road layout image used for training.

A key difference between our task and the original one in \cite{lu_image-graph-image_2020} is that, we aim to handle real-world road layout data, instead of parsing simple synthetic shapes into graphs. This results in new challenges for properly training the network. As we demonstrate in Section~\ref{sect_exp} as ablation study, the original naive training, \ie, feeding only the real-world road layout data for auto-encoding, is not sufficient to encourage the convergence of learning (please see the large standard deviation in the first row of Table~\ref{tab_ablation_trainingdata}). This is because our differentiable drawing module generates coarse images using synthetic line segments. It is sufficient to reconstruct the structural road layout information, but is not able to generate non-structural real-world road patterns, \eg, minor zigzags, even with the image refinement module applied. As a result, the pixel-wise SSIM loss cannot provide adequate supervision on the structural layout level, instead, the non-structural real-world patterns will also be reflected into SSIM loss, which is not desired and inhibits the training. Thus, additional contribution needs to be made to enable the training on real-world data.

As a solution, we propose to train the auto-encoder with two types of data simultaneously: \textit{real-world} road layouts and \textit{synthetic} road layouts. The synthetic data, as detailed in Section~\ref{sect_exp_datasets}, is randomly generated to simulate the common road pattern, \ie, crossroad. Thus, for each mini-batch $I_{\text{BEV}}$ used for training, it contains two sub-batches from two datasets: $I_{\text{BEV}} = [I_{\text{BEV\_real-world}}, I_{\text{BEV\_synthetic}}]$. The synthetic dataset serves as a bridge to merge the domain gap between the structural patterns in the drawing module and the real-world data containing non-structural patterns. The effect of different proportions of two datasets is provided by an ablation study in Section~\ref{sect_exp_datasets}.

\section{Experiments}
\label{sect_exp}

We perform the following experiments to validate the functionality and performance of our approach, with the experimental settings detailed first and the results discussed in Section~\ref{sect_results}:
\begin{itemize}
	\item We evaluate our approach on real-world Argoverse dataset, and compare the performance against a fully-supervised baseline;
	\item We perform the ablation study of different proportions of real-world and synthetic datasets used during training;
	\item We demonstrate a combination of our approach with a state-of-the-art monocular road layout estimation algorithm. This results in an end-to-end pipeline for high-level road layout parsing from front-view RGB images.
	\item We also showcase the capability of our approach to handle large-scale road layouts with minor extra processing.
\end{itemize}

\subsection{Datasets}
\label{sect_exp_datasets}

\begin{figure}[!tbp]
	\centering
	\includegraphics[width=\linewidth]{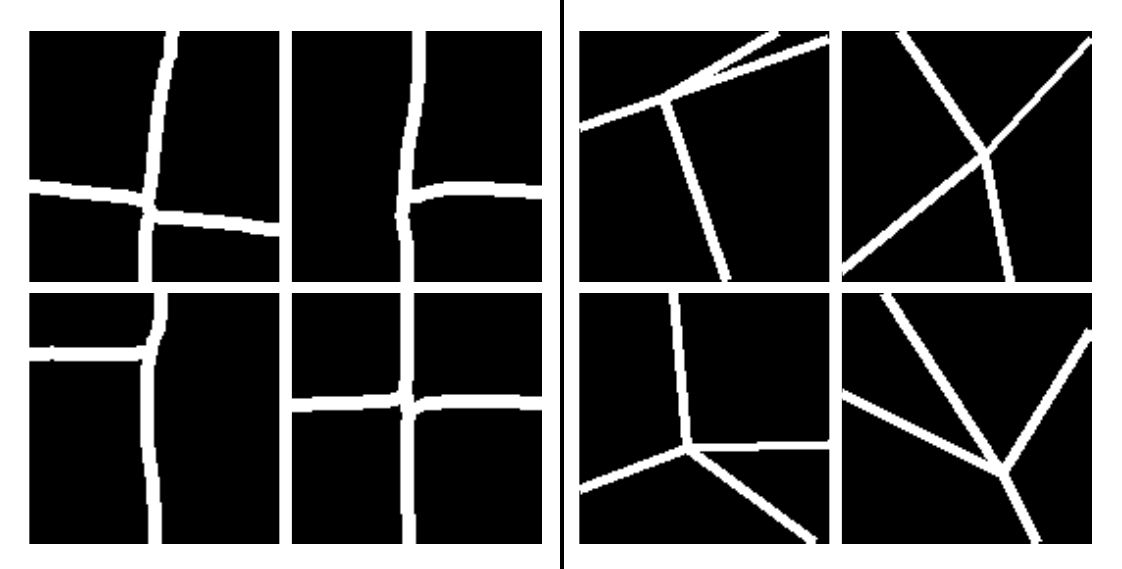}
	\caption{Visualizations of the two datasets. Left: real-world Argoverse dataset; Right: synthetic dataset.}
	\label{fig_dataset_samples}
\end{figure}

We use two datasets, \ie, real-world Argoverse and synthetic, to enable the graph learning of road layouts and evaluate the performance of different approaches. Some samples are illustrated in Figure~\ref{fig_dataset_samples}.

\begin{figure}[!tbp]
	\centering
	\includegraphics[width=0.9\linewidth]{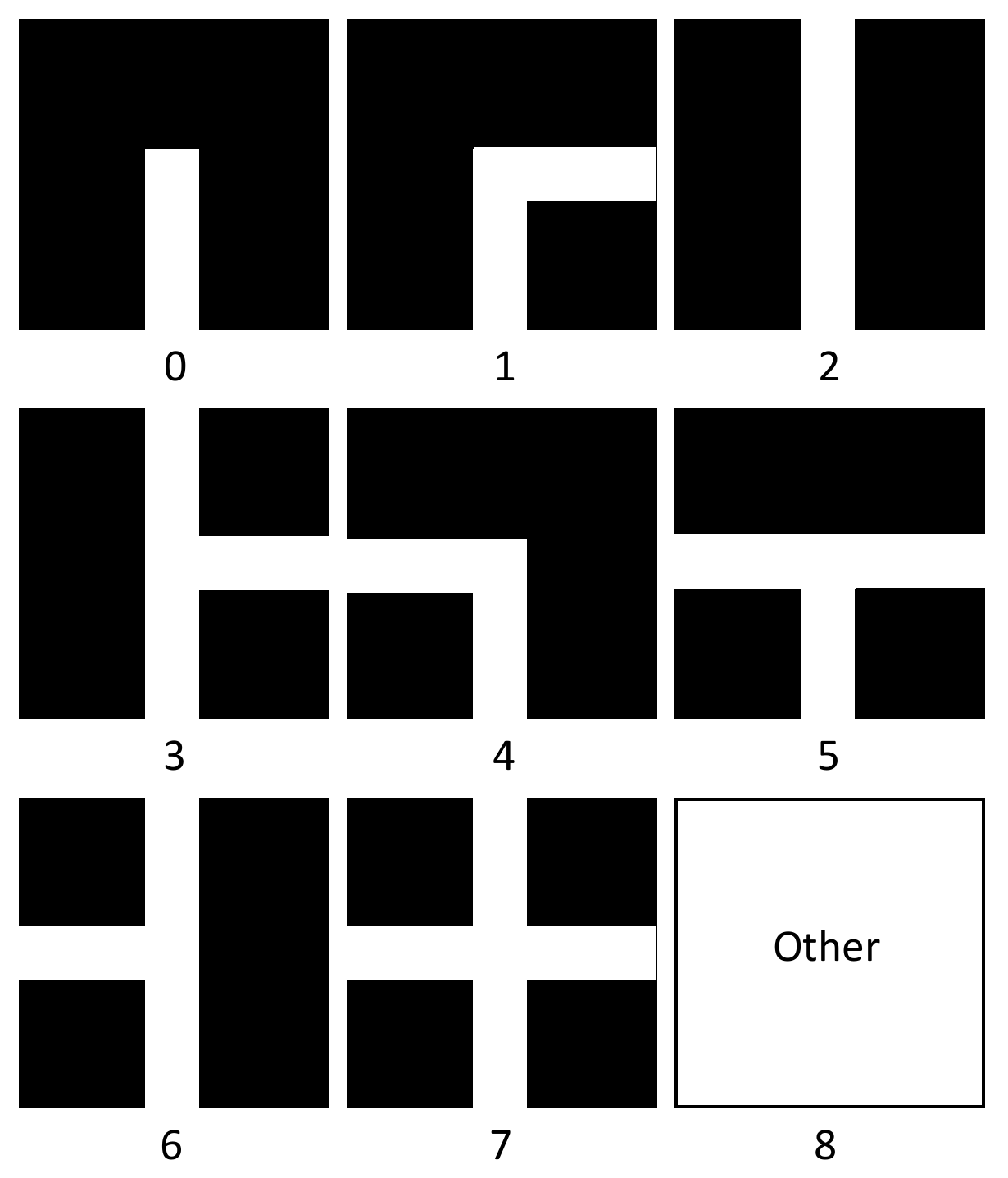}
	\caption{The visualization of 9 potential topologies, based on the assumption that ego-vehicle is located at the bottom of the BEV local road layouts. These labels are used to represent the class of road layouts for calculating the road topology classification accuracy metric. 500 samples from Argoverse dataset \cite{chang_argoverse_2019} are manually annotated for evaluating the topology classification accuracy given the predicted road graphs.}
	\label{fig_road_topologies}
\end{figure}

\textbf{Argoverse} This dataset \cite{chang_argoverse_2019} originally provides a collection of road layouts in BEV as ground truth for the task of front-view RGB image parsing \cite{mani_monolayout_2020}. In our experiments, we directly take these road layout images and pre-process them with thinning operation to eliminate the factor of varying road width. The dataset settings and train/validation split are aligned with the settings in \cite{mani_monolayout_2020}. Each image has 128 $\times$ 128 pixels, which correspond to a rectangular region of 40m $\times$ 40m. To fairly evaluate the performance of our task, ground truth of the test set is required. However, real-world road layouts contain various ambiguous non-structural patterns, thus it is not possible to clearly annotate a single set of road joints for most samples. Instead, we opt to assign a high-level topology label for each sample, according to 9 pre-defined layout scenarios, as illustrated in Figure~\ref{fig_road_topologies}. These labels are used for evaluation with the road topology classification accuracy, which is detailed in Section~\ref{sect_exp_metrics}. Thus only for validation, we manually annotate 500 samples in Argoverse test set in \cite{mani_monolayout_2020}, which are shared with the community.

\textbf{Synthetic road layouts} Given the human prior knowledge of road, we also generate a synthetic dataset with each sample representing a random road layout, which is used to train the proposed auto-encoder together with the real-world Argoverse dataset. The motivation of including this synthetic dataset for training is detailed in Section~\ref{sect_method_training}. To generate each sample, 5 nodes are randomly sampled, where 4 are located on 4 edges of an image and the remaining one is near the center of that image. Since this dataset is automatically generated given the corresponding random graphs without any real-world patterns, the precise ground truth graphs are available for graph-level performance evaluation, \eg, using triplet matching score. In a way, this efficient approach allows transferring human knowledge into a form that can be transferred to the deep data-driven model.

\subsection{Metrics}
\label{sect_exp_metrics}

We use two metrics to evaluate the quality of learned graphs describing road layouts: 1) triplet matching score, and 2) road topology classification accuracy. 

\textbf{Triplet matching score} This metric is derived from the scene graph generation task \cite{lu_visual_2016, yang_graph_2018}. Similarly as in \cite{lu_image-graph-image_2020}, where each line segment is considered as a triplet of two connected nodes, in our task, the road layouts could also be represented as a set of triplets of road joint pairs with their corresponding connectivity. To evaluate, we first perform the matching of ground truth and prediction triplets, and then calculate the precision, recall, and F1-score. Please note that this metric is used only on synthetic dataset since graph-level ground truth is not available for the real-world Argoverse dataset, as stated in Section~\ref{sect_exp_datasets}. Furthermore, even with ground truth available, it is not fair to evaluate on real-world data: during inference, extra joints may be detected in the middle of a zig-zag road segment and result in redundancy of road segments. These effects are expected and normal on real-world data, and will significantly reduce the triplet matching score. However, it does not imply the topological parsing and understanding are actually degraded, as long as the structural information is still captured.

\textbf{Road topology classification accuracy} To evaluate the performance on real-world datasets, we opt to classify the topology of each road layout image and evaluate the classification accuracy. Inspired by \cite{ballardini_online_2017}, we pre-define 9 types of topologies, depending on the accessibility of the left, front, and right borders of the BEV road layout image patches, as visualized in Figure~\ref{fig_road_topologies}. These definitions are based on the practical assumption that the ego-vehicle is located at the bottom of each image, which also holds in the Argoverse dataset. For each predicted graph, we use a simple rule-based algorithm to decide the prediction label, and then compare it with the manually annotated ground truth label. The accuracy can therefore be calculated on the test set.

As mentioned before, given the same input, extra nodes and their adjacency can be predicted, which results in different graphs but depicts the same topology, thus we also report the \textit{average number of predicted nodes}, as an indication of redundancy in the graphs. We also calculate the \textit{reference} average number of nodes based on the ground truth labels and their minimal ideal graphs, which is 3.784 nodes for the Argoverse test set, and report the \textit{relative exceeding ratio} of predicted nodes compared with the ideal reference number. Note that given the same level of topology accuracy, a lower \textit{average number of predicted nodes} and \textit{relative exceeding ratio} are preferred. 

\subsection{Baseline}

Since we are the first to parse road layouts into structural graphs using a deep data-driven method, and evaluate it with layout topology classification accuracy, there is no existing approach that could be directly and fairly compared with. Thus, we set up our own strong yet comparable baseline: we directly deploy the same encoder used in our network, and train this encoder in a fully-supervised manner using ground truth graphs provided with node and adjacency matrices. However, in our task, the main data is from the real world and no ground truth is available. Thus, it is not possible to train the baseline using real-world data, and we only train it with the aforementioned synthetic road layout dataset. The conventional loss functions, \eg, binary cross entropy, are applied to supervise the learning of encoder's node attention and relation classification modules. Please note that, strictly speaking it is still not fair to compare two methods, because the conditions for supervision and used datasets for training are different. However, we believe the comparison could fairly reveal the advantages and disadvantages of the proposed approach.

\begin{figure*}[!tbp]
	\centering
	\begin{subfigure}[t]{.15\textwidth}
		\centering
		\includegraphics[width=\linewidth]{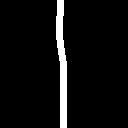}\\
		\vspace{2pt}
		\includegraphics[width=\linewidth]{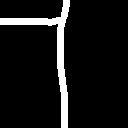}\\
		\vspace{2pt}
		\includegraphics[width=\linewidth]{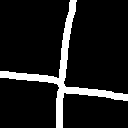}\\
		\caption{\centering Input}
	\end{subfigure}
	\begin{subfigure}[t]{.15\textwidth}
		\centering
		\includegraphics[width=\linewidth]{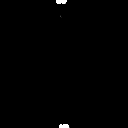}\\
		\vspace{2pt}
		\includegraphics[width=\linewidth]{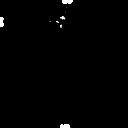}\\
		\vspace{2pt}
		\includegraphics[width=\linewidth]{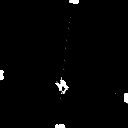}\\
		\caption{\centering Baseline node attention}
	\end{subfigure}
	\begin{subfigure}[t]{.15\textwidth}
		\centering
		\includegraphics[width=\linewidth]{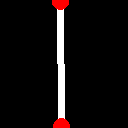}\\
		\vspace{2pt}
		\includegraphics[width=\linewidth]{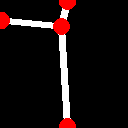}\\
		\vspace{2pt}
		\includegraphics[width=\linewidth]{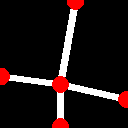}\\
		\caption{\centering Baseline visualized graph}
	\end{subfigure}
	\begin{subfigure}[t]{.15\textwidth}
		\centering
		\includegraphics[width=\linewidth]{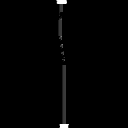}\\
		\vspace{2pt}
		\includegraphics[width=\linewidth]{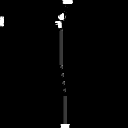}\\
		\vspace{2pt}
		\includegraphics[width=\linewidth]{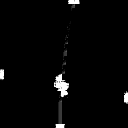}\\
		\caption{\centering Ours node attention}
	\end{subfigure}
	\begin{subfigure}[t]{.15\textwidth}
		\centering
		\includegraphics[width=\linewidth]{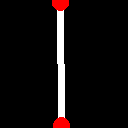}\\
		\vspace{2pt}
		\includegraphics[width=\linewidth]{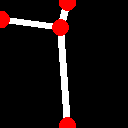}\\
		\vspace{2pt}
		\includegraphics[width=\linewidth]{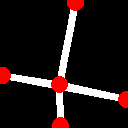}\\
		\caption{\centering Ours visualized graph}
	\end{subfigure}
	\caption{Qualitative results of our unsupervised approach and the baseline trained in a fully-supervised manner on synthetic road dataset. In the visualized graphs, red dots represent detected road joints, and white line segments represent their connectivity.}
	\label{fig_result_main}
\end{figure*}

\begin{table*}
	\renewcommand{\arraystretch}{1}
	\caption{Quantitative comparison of our unsupervised approach and the baseline trained in a fully-supervised manner on synthetic road layouts.}
	\begin{center}
		\begin{tabular}{c||cc|cccc}
			\hline
			\multirow{2}{*}{method} & \multirow{2}{*}{supervision} & \multirow{2}{*}{trained datasets} & triplet matching F1 score &  topology accuracy & avg. number of nodes & exceeding ratio\\
			&  &  & on synthetic dataset &  on Argoverse dataset &  on Argoverse dataset & on Argoverse dataset\\
			\hline
			\hline
			ours & none & both & 92.6$\pm$0.9 & \textbf{80.3}$\pm$2.1 & \textbf{4.2}$\pm$0.1 & \textbf{11.0\%}\\
			baseline & fully available & only synthetic & \textbf{94.7}$\pm$1.2 & 74.8$\pm$10.5 & 4.4$\pm$0.2 & 16.3\%\\
			\hline
		\end{tabular}
	\end{center}
	\label{tab_exp}
\end{table*}

\subsection{Implementation details}

To train our auto-encoder, we use Adam \cite{kingma_adam:_2014} optimizer with $\beta_1 = 0.6$, $\beta_2 = 0.9$, and initial learning being 0.0005. For each run, we train our network with batch size 32 for 12k iterations. Within each batch of data, the proportion of real-world and synthetic data could be changed, as described in Section~\ref{sect_method_training}. For the fully-supervised baseline, we find that the sub-task for the node attention prediction is particularly easy to train and overfit, and the supervision quality for the adjacency matrix is dependent on the quality of the predicted nodes, which is used for generating the temporary ground truth on the fly. Thus, for the first two epochs we train the node attention module only, and later we add the supervision loss for adjacency classification together with the node attention loss. All the experiments are performed using PyTorch \cite{paszke_automatic_2017}.

Please note that our learning objective, \ie, image reconstruction loss, is not directly correlated to the evaluation metrics, especially to road topology classification accuracy. This results in relatively large standard deviations for different training runs. Thus, for each experiment, we train the network with 5 random seeds and record the performance of the last 3 training epochs for each run. We therefore report mean values for each metric together with standard deviations in experiments.

\begin{table*}
	\renewcommand{\arraystretch}{1}
	\caption{Ablation of different ways to feed training data for our unsupervised approach.}
	\begin{center}
		\begin{tabular}{c||cccc}
			\hline
			\multirow{2}{*}{trained datasets} & triplet matching F1 score &  topology accuracy & avg. number of nodes & exceeding ratio\\
			&  on synthetic dataset &  on Argoverse dataset  &  on Argoverse dataset & on Argoverse dataset\\
			\hline
			\hline
			Argoverse  & 29.3$\pm$5.5 & 67.9$\pm$11.0 & 3.6$\pm$0.2 & -4.9\%\\
			25\% synthetic + 75\% Argoverse  & 84.1$\pm$4.2 & 75.2$\pm$4.4 & 3.8$\pm$0.1 & 4.2\%\\
			50\% synthetic + 50\% Argoverse  & 91.1$\pm$1.0 & 78.6$\pm$1.7 & 4.1$\pm$0.1 & 8.4\%\\
			75\% synthetic + 25\% Argoverse  & 92.6$\pm$0.9 & 80.3$\pm$2.1 & 4.2$\pm$0.1 & 11.0\%\\
			synthetic  & 93.8$\pm$0.8 & 83.7$\pm$1.8 & 6.6$\pm$0.6 & 74.4\%\\
			\hline
		\end{tabular}
	\end{center}
	\label{tab_ablation_trainingdata}
\end{table*}

\section{Results}
\label{sect_results}

\subsection{Road topology parsing}

\begin{figure*}[!tbp]
	\centering
	\begin{subfigure}[t]{.179766537\textwidth}
		\centering
		\includegraphics[width=\linewidth]{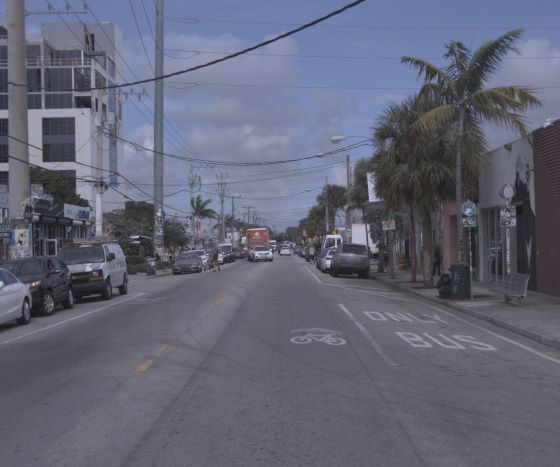}\\
		\vspace{2pt}
		\includegraphics[width=\linewidth]{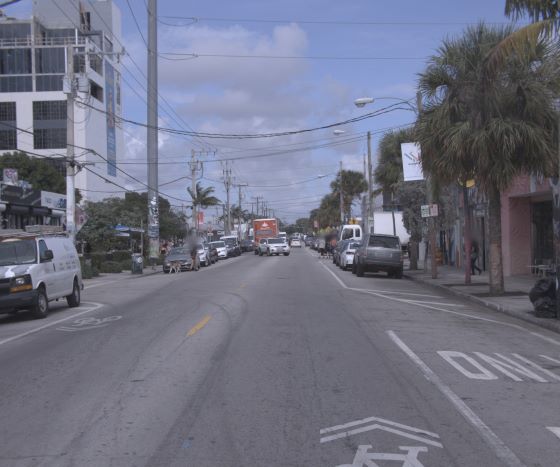}\\
		\vspace{2pt}
		\includegraphics[width=\linewidth]{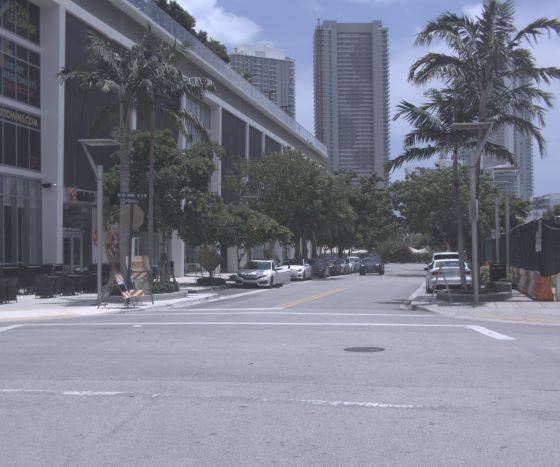}\\
		\caption{\centering Source RGB image}
	\end{subfigure}
	\begin{subfigure}[t]{.15\textwidth}
		\centering
		\includegraphics[width=\linewidth]{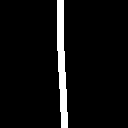}\\
		\vspace{2pt}
		\includegraphics[width=\linewidth]{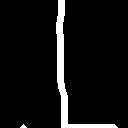}\\
		\vspace{2pt}
		\includegraphics[width=\linewidth]{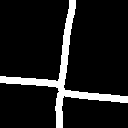}\\
		\caption{\centering State-of-the-art prediction}
	\end{subfigure}
	\begin{subfigure}[t]{.15\textwidth}
		\centering
		\includegraphics[width=\linewidth]{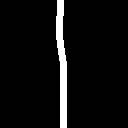}\\
		\vspace{2pt}
		\includegraphics[width=\linewidth]{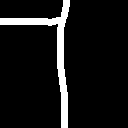}\\
		\vspace{2pt}
		\includegraphics[width=\linewidth]{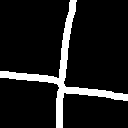}\\
		\caption{\centering Ground truth layout}
	\end{subfigure}
	\begin{subfigure}[t]{.15\textwidth}
		\centering
		\includegraphics[width=\linewidth]{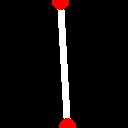}\\
		\vspace{2pt}
		\includegraphics[width=\linewidth]{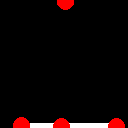}\\
		\vspace{2pt}
		\includegraphics[width=\linewidth]{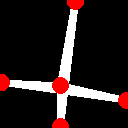}\\
		\caption{\centering Baseline visualized graph}
	\end{subfigure}
	\begin{subfigure}[t]{.15\textwidth}
		\centering
		\includegraphics[width=\linewidth]{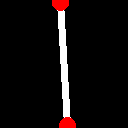}\\
		\vspace{2pt}
		\includegraphics[width=\linewidth]{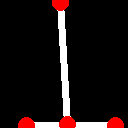}\\
		\vspace{2pt}
		\includegraphics[width=\linewidth]{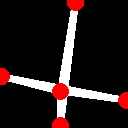}\\
		\caption{\centering Ours visualized graph}
	\end{subfigure}
	\caption{Quantitative comparison of our unsupervised approach and the fully-supervised baseline when using predictions of state-of-the-art method \cite{yang_projecting_2021} that maps front-view RGB images to road layout images. In the visualized graphs, red dots represent detected road joints, and white line segments represent their connectivity.}
	\label{fig_run_on_predictions}
\end{figure*}

In this experiment, we compare and discuss the main results of our auto-encoder and the fully-supervised baseline approach. The results are presented in Table~\ref{tab_exp} with some visualizations in Figure~\ref{fig_result_main}. The performances are evaluated on two datasets with different metrics, given the different types of ground truth available.

On the real-world Argoverse dataset, our approach achieves 80.3\% road layout topology classification accuracy, which outperforms the baseline by a large margin of 5.5\%. Furthermore, the significantly lower standard deviation (2.1\%) indicates the stronger stability of our approach over the baseline, which reaches 10.5\% of standard deviation. As for the average number of predicted nodes, our approach exhibits slightly better performance, \ie, 4.2 nodes (11.0\% exceeding ratio), which is 0.2 (5.3\% exceeding ratio) lower than the baseline. Together with the topology accuracy, one can conclude that our approach can understand the topology of road layouts using a simpler graph representation. It is understandable that our self-supervised approach outperforms the fully-supervised baseline on real-world Argoverse dataset, since the conditions of training data are different: we train our auto-encoder with both synthetic and real-world datasets; In contrast, the baseline can only be trained on the synthetic dataset, as the real-world data lacks ground truth for training. This explanation is also supported by the graph triplet matching F1 score evaluated on the synthetic dataset, where the baseline achieves 94.7\% and our approach is slightly lower (92.6\%). When the ground truth is available for training, the fully-supervised approach could naturally outperform the self-supervised counterpart.

Overall, both approaches are able to deliver acceptable performance on all metrics. However, since our approach can be trained without any external manual annotation, it is easier to be deployed in novel real-world scenarios with potential domain gaps, which is not possible for the fully-supervised baseline.

\subsection{Ablation on dual datasets training}

As indicated in Section~\ref{sect_method_training}, we opt to fuse synthetic and real-world data together during training of our auto-encoder. This ablation study, as listed in Table~\ref{tab_ablation_trainingdata}, shows the results when different proportions of two datasets are fused in each mini-batch. 

Five scenarios are tested in this experiment, and from Table~\ref{tab_ablation_trainingdata} a clear trend can be observed when increasing the proportion of synthetic data during training. When training with only real-world Argoverse dataset (first line), our approach is able to learn the graph, \ie, 67.9\% topology accuracy, but cannot provide stable predictions, as the standard deviation reaches 11.0\%. Furthermore, it cannot generalize on the synthetic dataset since it is never seen during training. By introducing more synthetic samples in each mini-batch, the triplet matching F1 score is significantly boosted, which is expected as the training data includes samples from the synthetic dataset. Interestingly, the topology classification accuracy on Argoverse dataset is also improved from 67.9\% to 83.7\% with much lower standard deviations, \ie, from 11.0\% to 1.8\%. This shows that introducing synthetic data could benefit the learning of real-world data, in terms of both performance and training stability, despite the facts that fewer real-world data samples are exposed during training, and the domain gap exists between synthetic and real-world data. 

On the other hand, the average number of nodes and the corresponding exceeding ratio also increase when introducing more synthetic data, especially when no real-world data is used (6.6$\pm$0.6 nodes and 74.4\% exceeding ratio), which is not preferred since extra nodes will complicate the predicted graph. Thus for our approach, we choose to take 75\% synthetic data and 25\% real-world Argoverse data for training, which leads to a relatively lower average number of nodes and higher graph classification accuracy.

\begin{table}
	\renewcommand{\arraystretch}{1}
	\caption{Quantitative comparison of our unsupervised approach and the baseline when using predictions of state-of-the-art method that maps front-view RGB images to road layouts.}
	\begin{center}
		\begin{tabular}{cc||cc}
			\hline
			used ground truth & method & topology accuracy & avg. nodes\\
			\hline
			\hline
			\multirow{2}{*}{original} & ours & \textbf{52.8}$\pm$1.9 & \textbf{5.0}$\pm$0.1 \\
			& baseline & 46.8$\pm$7.7 & 5.4$\pm$0.2\\
			\hline
			re-annotated & ours & \textbf{75.8}$\pm$1.3 & \textbf{5.0}$\pm$0.1 \\
			based on predictions& baseline & 64.9$\pm$8.0 & 5.4$\pm$0.2\\
			\hline			
		\end{tabular}
	\end{center}
	\label{tab_run_on_predictions}
\end{table}

\subsection{From front-view image to road layout graph}
\label{sect_resluts_full_pipeline}

\begin{figure*}[!tbp]
	\centering
	\begin{subfigure}[t]{0.32\textwidth}
		\centering
		\includegraphics[width=\linewidth]{imgs/large_scale/tracking3_UTM_cropped.pdf}\\
		\vspace{2pt}
	\end{subfigure}
	\begin{subfigure}[t]{.32\textwidth}
		\centering
		\includegraphics[width=\linewidth]{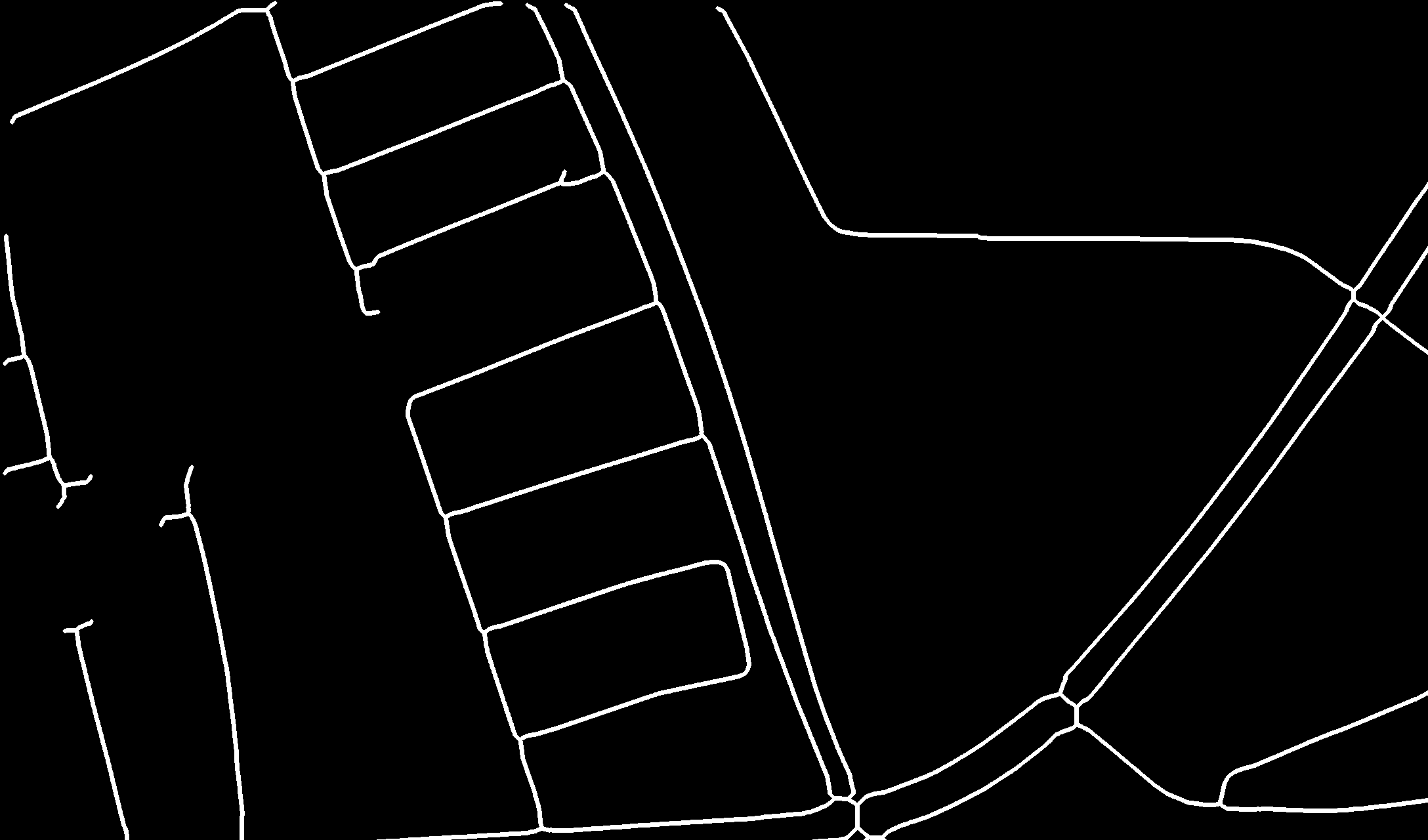}\\
		\vspace{2pt}
	\end{subfigure}
	\begin{subfigure}[t]{.32\textwidth}
		\centering
		\includegraphics[width=\linewidth]{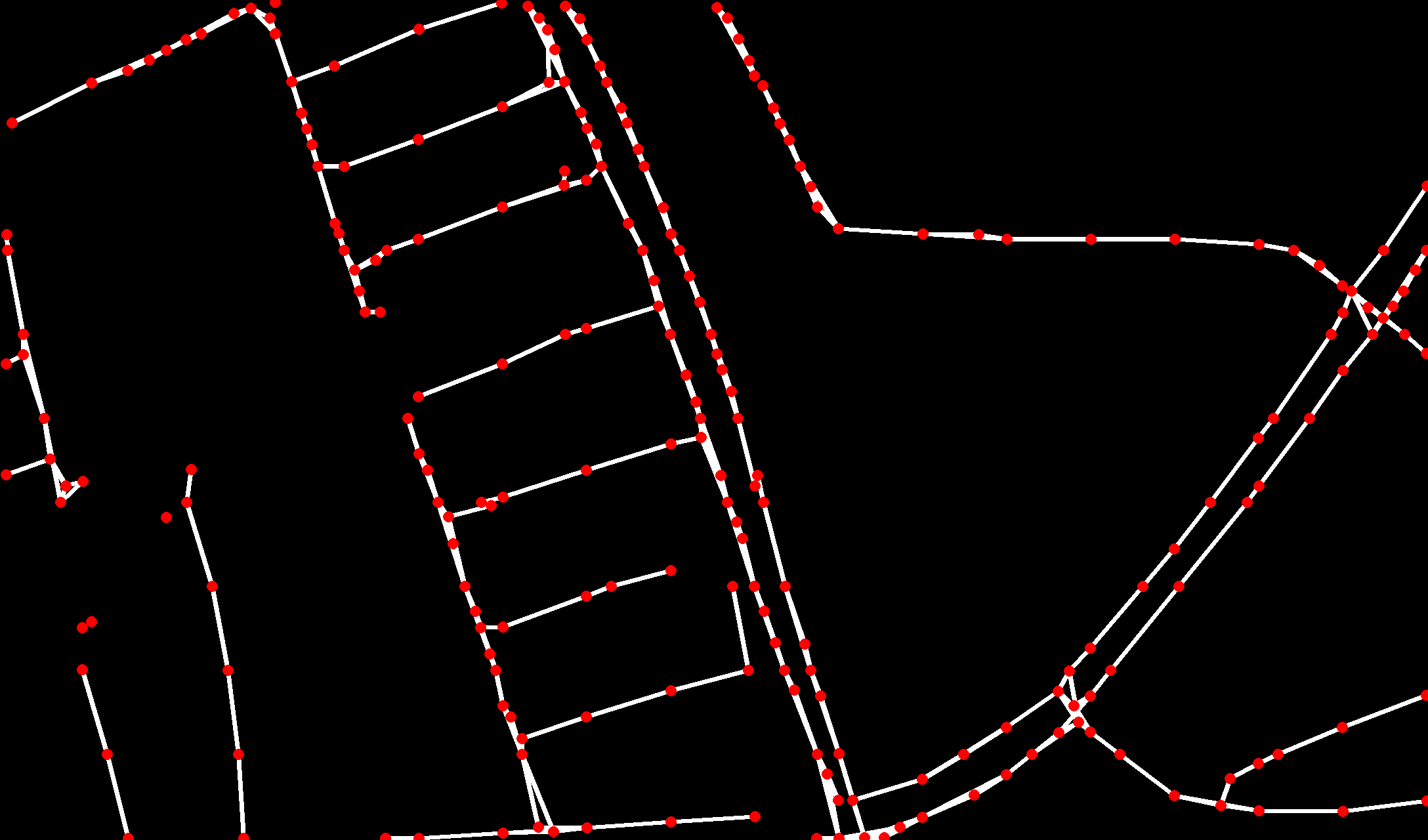}\\
		\vspace{2pt}
	\end{subfigure}
	\caption{Visualization of large scale road layout parsing. Left: RGB image for reference; Middle: input binary road layout image from \cite{mattyus_enhancing_2015}; Right: visualized road layout from the predicted large scale road layout graph, where red dots represent detected road joints, and white line segments represent their connectivity. Please zoom in for better visualization.}
	\label{fig_large_scale}
\end{figure*}

Recently, a series of research has been carried out on estimating the road layout in BEV from front-view RGB image \cite{lu_monocular_2019, mani_monolayout_2020, yang_projecting_2021}. To further showcase the potential applications of our approach, we hereby present the results on the  complete end-to-end task that is able to process raw front-view RGB images and predict the road layouts not only as images, but also as human-interpretable structural graph representations. This pipeline is formalized as a complete end-to-end task with two stages in Section~\ref{sect_method_pro_def}. It is potentially useful, since extra graph information could be used for a higher level of reasoning and planning of intelligent vehicles. To be specific, we apply our learned encoder directly on top of a state-of-the-art monocular road layout mapping network \cite{yang_projecting_2021}. Our network takes the estimated road layouts from \cite{yang_projecting_2021} as input and predicts graphs that describe the road scenes of the corresponding RGB images. Together with \cite{yang_auto-encoding_2019}, a pipeline is created that directly parses front-view RGB images into graphs.

The results are presented in Table~\ref{tab_run_on_predictions} and Figure~\ref{fig_run_on_predictions}. We evaluate the performance quantitatively on the same Argoverse dataset with two types of ground truth: 1) the original topology labels as indicated in Section~\ref{sect_exp_datasets} (upper two rows), and 2) the labels that are re-annotated based on the observation of state-of-the-art predictions (bottom two rows). When using the original ground truth, the performance degrades compared to the approaches that take ground truth as inputs (see Table~\ref{tab_exp}), yet still both are able to perform the task successfully. This performance drop is mainly due to some mis-predictions of road layout images from upstream monocular road layout mapping task \cite{yang_auto-encoding_2019}. This explanation is also supported by performance when using re-annotated ground truth based on the observation of state-of-the-art predictions, which is significantly higher in terms of topology accuracy. Thus, one can conclude that the limiting factor for this end-to-end task is the mapping from front-view images to BEV images, and not our graph encoder model.

Same as in the main experiment, our approach also outperforms the fully-supervised baseline by notable margins (6\% and 10.9\%, using two types of ground truth respectively), which indicates its generalizability and robustness against inputs with prediction errors.

\subsection{Large scale road layout parsing}

Our approach can easily be extended to process large-scale road layout images and parse them into a graph that contains a large set of road joints and an adjacency matrix describing their connectivity. To achieve this, we first break a large-scale road layout image into non-overlapping rectangular pieces and feed them into the learned encoder. Afterward, we merge the predicted sub-graphs into a single large-scale graph describing the complete road scene by simply applying node coordinate offsets and deleting redundant overlapping nodes. This application might potentially be useful for automated map-making from aerial images.

We use AerialKITTI dataset \cite{mattyus_enhancing_2015} to demonstrate this functionality. This dataset contains 21 large-scale binary road segmentation images, capturing the same area as in KITTI tracking dataset \cite{geiger_vision_2013}. We only provide some visualizations for this experiment since it is not possible to compute our metrics without having ground truth available, see Figure~\ref{fig_large_scale}. The predictions show that a large set of road joints and their connectivity are successfully extracted. This indicates that our neural network approach is able to achieve a high-level understanding of large-scale road layout using only annotation-free self-supervised learning.

\section{Conclusions}
\label{sect_conclusions}

In this work, we propose and demonstrate a self-supervised neural network approach to parse road layouts as graphs containing road joints and their connectivity. We achieve this by utilizing a carefully designed auto-encoder that learns to automatically regress graphs at its bottleneck during self-supervised learning on both real-world and synthetic datasets. During inference, the learned encoder takes any road layout image as input and predicts a corresponding human-interpretable graph. Our neural network, which is trained without any manual annotations, exhibits a decent performance compared to a strong fully-supervised baseline. Moreover, we show two potential use-cases of our approach in the context of intelligent vehicles and automated map-making. 

From an engineering perspective, currently model-based approach might still be preferred for these applications due to their explainable and controllable behavior. However these hand-crafted approaches are usually not scalable. Our approach has shown that such complex applications can alternatively be realized using self-supervised deep data-driven approaches that potentially offer much better scalability and also achieve higher accuracy and robustness.

\addtolength{\textheight}{-10cm}   

\bibliographystyle{./IEEEtran}
\bibliography{MyLibrary.bib,repo.bib}


\end{document}